\documentclass[10pt, journal]{IEEEtran}
\usepackage{multirow}
\usepackage{subfigure}
\usepackage{booktabs}
\usepackage{xspace} %
\usepackage{bbding} %
\usepackage{amsmath,amssymb,amsfonts}
\usepackage{algorithm,algpseudocode}
\usepackage{amsmath,bm}

\usepackage{graphicx}

\def\name{{Fed-EDKD}\xspace}
\def\attname{{C-GANs}\xspace}

\begin{document}

\title{Mitigating Cross-client GANs-based Attack in Federated Learning}

\author{\IEEEauthorblockN{Hong Huang\IEEEauthorrefmark{1},
~Xinyu Lei\IEEEauthorrefmark{2},
~Tao Xiang\IEEEauthorrefmark{1}
}
%~Chi-Yu Li\IEEEauthorrefmark{3}}
%Hongyu Huang\IEEEauthorrefmark{3},
%Tian Xie\IEEEauthorrefmark{1}}
\\
\IEEEauthorblockA{\IEEEauthorrefmark{1} College of Computer Science, Chongqing University,
Chongqing, China \\}
\IEEEauthorblockA{\IEEEauthorrefmark{2} Department of Computer Science, Michigan Technological
University, Houghton, USA \\}
%
%\IEEEauthorblockA{\IEEEauthorrefmark{3} Dept. of Computer Science, National Chiao Tung University, Taiwan \\}
%
Email: 0164478@cqu.edu.cn; xinyulei@mtu.edu; txiang@cqu.edu.cn \\% chiyuli@cs.nctu.edu.tw\\

\thanks{This paper is officially published in Multimedia Tools and Applications, 2023 \cite{huang2023mitigating}.
Corresponding author: Tao Xiang.}
}

\maketitle

%\title[ ]{Mitigating Cross-client GANs-based Attack in Federated Learning}


%\author[1]{\fnm{Hong} \sur{Huang}}\email{20164478@cqu.edu.cn}
%
%\author[2]{\fnm{Xinyu} \sur{Lei}}\email{xinyulei@mtu.edu}
%
%\author*[1]{\fnm{Tao} \sur{Xiang}}\email{txiang@cqu.edu.cn}

%\affil*[1]{\orgdiv{College of Computer Science}, \orgname{Chongqing University}, \orgaddress{ \city{Chongqing}, \country{China}}}
%
%\affil[2]{\orgdiv{Department of Computer Science}, \orgname{Michigan Technological University}, \orgaddress{\city{Houghton}, \country{United States of America}}}


\begin{abstract}

Machine learning makes multimedia data (e.g., images) more attractive, however, multimedia data is usually distributed and privacy sensitive.
Multiple distributed multimedia clients can resort to federated learning (FL) to jointly learn a global shared model without requiring to share their private samples with any third-party entities.
%w tha
In this paper, we show that FL suffers from the cross-client generative adversarial networks (GANs)-based (\attname) attack, in which a malicious client (i.e., adversary) can reconstruct samples with the same distribution as the training samples from other clients (i.e., victims).
%
%\textcolor{blue}{
Since a benign client's data can be leaked to the adversary, this attack brings the risk of local data leakage for clients in many security-critical FL applications.
Thus, we propose \name (i.e., Federated Ensemble Data-free Knowledge Distillation) technique to improve the current popular FL schemes to resist \attname attack.
In \name, each client submits a local model to the server for obtaining an ensemble global model.
Then, to avoid model expansion, \name adopts data-free knowledge distillation techniques to transfer knowledge from the ensemble global model to a compressed model.
By this way, \name reduces the adversary's control capability over the global model, so \name can effectively mitigate \attname attack.
Finally, the experimental results demonstrate that \name significantly mitigates \attname attack while only incurring a slight accuracy degradation of FL. 
\end{abstract}



\section{Introduction}\label{sec1}

Nowadays, machine learning (ML) technology is extensively applied to various fields such as computer vision, natural language processing, and speech recognition, etc.
A good ML model requires a large amount of multimedia data (e.g., images, text, audio, etc.) for training to ensure high performance.
In practice, a large training dataset  is usually collected by a group of distributed multimedia devices, such as computers, laptops, smartphones and other electronic sensors.
Meanwhile, these multimedia devices are generating large amounts of data all the time.
The traditional approach is to let many distributed multimedia clients send their local datasets to a centralized server for training.
However, the clients' local datasets may contain sensitive data (e.g., personal biometric data, location data, commercial valuable data, etc.) so that the clients may not want to share them with others.
Therefore, how to protect the privacy of the clients has become an urgent concern.

To address the security problems, federated learning (FL) technique is proposed, which enables a group of distributed clients to collaboratively train an ML model with the help of a central server for coordination.
In FL, each client only shares local model updates, and does not need to share its local dataset with others, and therefore, their local data privacy is preserved.
Therefore, FL applications are widely used in various distributed, privacy sensitive scenarios \cite{liu2022federated,peyvandi2022privacy}.
Current popular federated learning techniques (e.g., FedAvg \cite{mcmahan2017communication}, FedProx\cite{li2020federated}, SCAFFOLD\cite{karimireddy2020scaffold}, and MOON\cite{li2021model}) require multiple interactions between clients and server to train a global ML model.
After sufficient training, a global ML model can be obtained.
A similar technique is collaborative learning \cite{shokri2015privacy}, which can be classified as sequential collaborative learning and parallel collaborative learning. 
In sequential collaborative learning, global ML model is trained on each client in sequential order.
Usually, parallel collaborative learning is also named FL.
In FL, clients' local model updates are sent to the central server where these updates are averaged to update the global model.

A recent study shows that sequential collaborative learning is vulnerable to GANs-based data reconstruction attacks \cite{hitaj2017deep}.
In this attack, a malicious client (i.e., the adversary) can apply adversarial influence to the learning process and utilize the shared model as the discriminator to secretly train GANs for reconstructing samples with the same distribution as victim's private samples.
While the adversarial influence applied may become trivial after the average update in FL.
Because of the difference between the sequential collaborative learning and the parallel collaborative learning (i.e., FL), we aim to test the performance of this attack in FL.
In this paper, inspired by the above attack, we design a \textbf{C}ross-client \textbf{GANs}-based (\attname) attack in FL.
In \attname attack, a malicious client (i.e., adversary) does not own the data of a target class.
Our experiments show that \attname attack enables the adversary to reconstruct samples with the same distribution as the training samples of the target class, so the adversary can easily learn the private information of other clients' local data.
%
%For example, a malicious client may not have data belonging to a certain class, the client can use C-GANs attack to reconstruct data from other clients. 
%
Recall that FL is designed to protect each client's local dataset from other entities.
%
%\textcolor{blue}{
However, \attname attack brings the risk of local data leakage for clients in many security-critical FL applications.
For example, a malicious client can reconstruct the face image belonging to other clients in FL.
Since \attname attack breaks the validity of the traditional FL scheme, how to mitigate \attname attack is identified as one of the most fundamental problems in FL. 
Thus, it is imperative to revise the traditional FL scheme to mitigate \attname attack.

In \attname attack, the adversary trains GANs by utilizing the shared global model and does not need to access other clients' gradients.
Therefore, based on secure multi-party computing and homomorphic encryption, the known schemes for protecting gradients in FL are ineffective for against \attname attack.
To mitigate GANs-based attack \cite{hitaj2017deep} in collaborative learning, two solutions \cite{chen2020secure,luo2020exploiting} have been proposed, however, both of them suffer from some limitations. 
Chen et al. \cite{chen2020secure} proposed a solution to isolate the participants from the global model parameters during FL training.
Its major limitation is to assume the existence of an additional trust third party.
In \cite{luo2020exploiting}, GANs are trained to generate fake samples participating in collaborative learning to protect the original training data.
There are two limitations in this approach, first, it leads to relatively high performance degradation, second, training GANs costs intensive computational overhead on clients.

In this paper, we improve the traditional FL scheme and further propose \textbf{Fed}erated \textbf{E}nsemble \textbf{D}ata-free \textbf{K}nowledge \textbf{D}istillation scheme (named \name) to mitigate \attname attack.
In \name design, we adopt the ensemble learning technique \cite{opitz1999popular}, which enables each client to submit a local learned ML model to the central server.
The server can combine the received local models to obtain a global model with better predictive performance.
By using ensemble learning in federated learning, a malicious client can only manipulate its local ML model, so the malicious client's control capability over the global ML model is significantly reduced.
Therefore, \name can effectively mitigate the \attname attack.

There are three major technical challenges in \name to be addressed.
First, general ensemble federated learning only requires single client-server iteration \cite{cao2021provably}, so \name may lead to a large performance degradation of global model.
To minimize the performance degradation, \name adopts multiple iterations for training a global model with high predictive performance.
After the server combines the global model using ensemble strategy, the server distributes it to each client.
In the next iteration, each client can use the received global model to do local training and submit it to the server for ensemble learning again.
After multiple iterations, the predictive performance of the trained global model by \name can be significantly increased.
Second, it is challenging to handle the global model expansion issue.
If we use the above multiple-iteration strategy, then the size of global model is expanded in each iteration (because the standalone ensemble global model has larger size than each local model).
To solve this challenge, \name exploits knowledge distillation (KD) technique, which transfers the knowledge from the large ensemble model (teacher model) to a smaller compressed student model.
Therefore, the model expansion issue can be solved.
Third, it is challenging to achieve highly precise KD when the training data is absent on the central server.
In FL, the server is prohibited to access the clients' training data, so it is hard to perform highly precise KD.
To tackle this challenge, \name employs the data-free KD technique.
A generator is trained to produce the imitated training data, which can facilitate the knowledge transfer from the teacher model to the student model.
Hence, the highly precise KD can be accomplished by the server without accessing real training data of clients.

In summary, this paper makes the following three main contributions.
%\vspace{-0.35in}
\begin{itemize}[]
	
	\item[$\bullet$] We experimentally test the \attname attack performance in some state-of-the-art FL schemes, and discover conditions under which the attack has good or poor performance. 
	
	\item[$\bullet$] We design \name to effectively mitigate \attname attack. 
	Besides, \name combines the ensemble learning and data-free knowledge distillation techniques to ensure the high performance of \name.
	
	\item[$\bullet$] \name enjoys two additional immediate benefits.
	First, it does not require revising any protocol from the client-side, and only server-side protocol update is applied. 
	Second, \name can achieve higher performance without sacrificing security compared with several defense strategies.
\end{itemize}

%\textcolor{blue}{
The rest of this paper is organized as follows.
Some related works are reviewed in Section \ref{sec:relate-work}.
Section \ref{sec:Preliminary} introduces some related preliminaries.
\attname attack is described in detail in Section \ref{sec:attack}.
Section \ref{sec:system} overviews the proposed \name.
Section \ref{sec:data-free} introduces the employed data-free knowledge distillation scheme.
% 
%The overall training process of \name is summarized in Section \ref{sec:algorithm}.
% %
Section \ref{sec:Experiment} exhibits experimental results, followed by Section \ref{sec:discuss}, which discusses the proposed \name.
Section \ref{sec:conclusions} concludes this paper.
To support reproducibility, some experiment details can be found in Appendix \ref{sect:appendix}.

\section{Related Work}\label{sec:relate-work}
In this section, we review the related work.

\subsection{Privacy Leakage in Collaborative Learning}

Although clients only share their gradient information in collaborative learning, it is still vulnerable to various attacks including data reconstruction attack, membership inference attack, backdoor attack, and attribute inference attack.
Data reconstruction attack aims to reconstruct images of the same distribution from participating clients.
Wang et al. \cite{wang2019beyond} present a method that assumes a malicious server and incorporates GANs with a multi-task discriminator to infer class representatives of a certain client.
But this method only works under certain circumstances where the reconstructed data is mostly homogeneous across clients and shared updates are plaintext.
Fredrikson et al. \cite{fredrikson2015model} proposed a method, which can roughly reconstruct a victim's face image from a block model with only given a name.
Although it works well for MLP, it is not ideal for CNNs.
Furthermore, Zhu et al. \cite{zhu2019deep} proposed DLG to reconstruct training samples by optimizing the random input to generate the same gradients for a specific client.
iDLG \cite{zhao2020idlg} improves the efficiency of DLG.
However, these two methods are effective only if the shared gradients are plaintext and become invalid when the batch consists of many samples.
The membership inference attack is to infer whether a given sample is in the training dataset or not.
Shokri et al. \cite{shokri2017membership} proposed the membership inference attack against a well-trained model by utilizing the differences in the target model’s predictions on trained inputs and that not trained.
Studies \cite{gu2022cs,nasr2019comprehensive} extend the membership inference attack to federated learning.
Backdoor attack \cite{bagdasaryan2020backdoor} utilizes the well-designed gradients (sent to the server) to modify the global model in the last round, so the adversary can insert backdoor functionality into the joint trained model.
Attribute inference attack \cite{melis2019exploiting} utilizes the shared model updates to infer sensitive attributes of training data. (e.g., specific locations).

\subsection{Privacy-Preserving Collaborative Learning}

Existing privacy-preserving collaborative learning methods can be categorized into four types : differential privacy (DP), secure multi-party computing (SMC), homomorphic encryption (HE) and robust aggregation.
DP introduces a certain amount of randomness or noise in the output to mask the user's influence on the output \cite{dwork2006differential}.
Shokri et al. \cite{shokri2015privacy} first proposed collaborative deep learning at realistic distribution and realized DP on it.
Recently, Robin et al. \cite{geyer2017differentially} proposed client-level differential privacy-preserving federated optimization to further protect the information of clients.
However, these DP-based methods make the train process are difficult to converge and realize a desirable privacy-performance tradeoff.
In addition, SMC has also been applied to protect the shared updates in privacy-preserving collaborative learning.
The SMC \cite{yao1986generate} security model involved multiple parties and guaranteed complete zero-knowledge, that is, each party knew nothing except input and output.
The authors \cite{mohassel2017secureml} utilized SMC to design a secure ML framework with two servers and semi-honest assumptions.
Besides, SMC technique is used in \cite{kilbertus2018blind} for model training and verification without revealing sensitive attributes of users.
Homomorphic encryption is a form of encryption that enables users to perform computations on their encrypted data without first decrypting it and the results of decryption produce the same output as if the operation are performed on the unencrypted data.
Recently, Jing et al. \cite{ma2022privacy} proposed a multi-key homomorphic encryption protocol to protect the shared gradients in FL.
Truex et al. \cite{truex2019hybrid} proposed a method combining DP and SMC that enables FL to reduce the growth of noise injection as the number of clients increases without sacrificing privacy while maintaining a pre-defined rate of trust.
However, both these SMC-based and HE-based methods will cost unbearable computational and communication overhead.
Robust aggregation improves the resilience of aggregation algorithms by carefully choosing the local model updates for aggregation, or balances updates from malicious clients by adding noise to the aggregation model \cite{sun2021decentralized,blanchard2017machine}.
On the other hand, various anomaly detection methods have been proposed to identify malicious clients' local model updates \cite{nguyen2020poisoning,tolpegin2020data}.

\section{Preliminaries}\label{sec:Preliminary}

\noindent
\textbf{Federated Learning.}
Federated learning is a distributed ML technique that enables training an algorithm across multiple distributed clients holding local dataset, without exchanging them.
There are two types of entities (a central server and multiple clients) in the FL system model.
At each iteration, clients download the global model from the central server, and then train it locally using its local dataset according to the initially agreed upon common algorithm.
Then, these local model updates (i.e., the difference between the local model and the global model) are sent to the central server where these updates are averaged to update the global model, and then the updated global model is redistributed to the clients for the next iteration.
The above processes are performed in multiple iterations and terminate until the ML model is well-trained.

\noindent
\textbf{Generative Adversarial Networks.}
GANs are first proposed by Goodfellow et al. \cite{goodfellow2014generative}.
GANs are a method of unsupervised learning, where two neural networks learn by playing against each other.
GANs consist of a generator model ($G$) and a discriminator model ($D$).
For $G$ training, $G$ takes random noise variable $\boldsymbol{z}$  from a prior distribution (e.g., Gaussian or uniform distribution) as the input, and then $G$ is trained to generate fake samples simulating the real samples of training data.
For $D$ training, $D$ takes the fake samples generated by $G$ and real training sample $\boldsymbol{x}$ as the input, and then $D$ is trained to distinguish the real samples and fake samples as much as possible.
The above $G$ training and $D$ training are performed multiple rounds interactively.
GANs' training process can be mathematically expressed as
\begin{equation}
	\begin{aligned}
		\min\limits_{G}\max \limits_{D} V(D,G) = \mathbb{E}_{\boldsymbol{x} \sim p_{data}(\boldsymbol{x})} [\log D(\boldsymbol{x})] + \\ \mathbb{E}_{\boldsymbol{z} \sim p_{\boldsymbol{z}}(\boldsymbol{z})} [\log(1-D(G(\boldsymbol{z})))],
	\end{aligned}
\end{equation}
where $p_{data}(\boldsymbol{x})$ denotes the real distribution and $p_{\boldsymbol{z}}(\boldsymbol{z})$ denotes the prior distribution.

\section{Cross-client GANs-based Attack}\label{sec:attack}

\begin{figure*}[h]
	\centering
	\includegraphics[width=1.9\columnwidth]{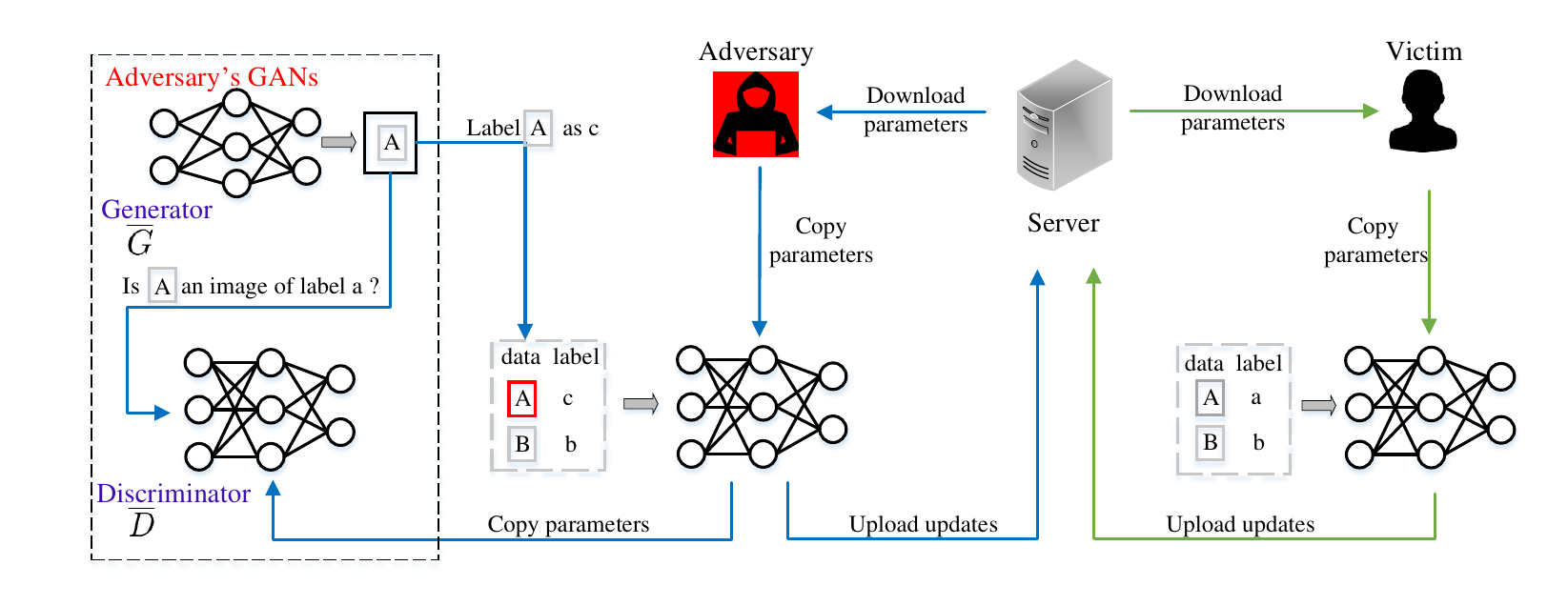}
	\caption{\centering{\attname attack on federated learning.}}
	\label{fig:CCG}
\end{figure*}

Hitaj et al. \cite{hitaj2017deep} developed a GANs-based attack in the sequential collaborative learning.
Given the similarity between the sequential collaborative learning and the parallel collaborative learning (i.e., FL), we intend to test the cross-client GANs-based (\attname) attack performance in FL.
In the following, we first introduce the threat model and some details of \attname attack.
The experimental evaluation for \attname attack is presented in Section \ref{sec:Experiment}.

\subsection{Threat Model}
The adversary is assumed to be one of the clients in FL.
The central server follows the FL protocol.
We consider the adversary does not own the data of a target label.
The goal of the adversary is to reconstruct samples with the same distribution as the target label's samples as accurately as possible.
Therefore, the victim(s) may be one or multiple clients who own the data with the target label.
Besides, the adversary can adaptively upload specially-crafted updates to trick other clients into leaking more information on their local data.
The leaked information can help the adversary to reconstruct the target label's samples.
Note that the adversary's attack is performed without changing the FL process, so this attack is carried out stealthily.

\subsection{\attname Attack}\label{atk-design}

Fig. \ref{fig:CCG} shows how \attname attack works in FL.
There are two clients, i.e., one is the adversary and the other is the victim.
We assume that the adversary owns the data that has two labels $\{b, c\}$ and the victim owns the data that has two labels $\{a, b\}$.
In \attname attack, the adversary aims to reconstruct samples with the same distribution as the victim's samples (with target label $a$) as much as possible.
Note that the samples with label $a$ are only owned by the victim.
The most frequently used protocol, FedAvg, is adopted.

To perform \attname attack, the adversary designs a local generator $\overline{G}$ and exploits the shared model in FL as the discriminator $\overline{D}$.
The generator $\overline{G}$ and the discriminator $\overline{D}$ form the adversary's GANs.
As shown in Fig. \ref{fig:CCG}, in one round of FL training, the adversary first trains $\overline{G}$.
Then, the trained $\overline{G}$ is used to generate a group of fake samples of class $a$ (from the victim's dataset).
Next, the adversary labels the generated fake samples as class $c$.
The labeled fake samples are injected into the adversary's local dataset.
The samples with label $c$ can be treated as fake samples and the samples with label $a$ can be treated as the real samples.
Therefore, the shared model can be treated as $\overline{D}$ trained by both fake and real samples.
During the FL, the adversary and the victim jointly train $\overline{D}$, while $\overline{G}$ is trained only by the adversary.
After multiple iterative training, $\overline{G}$ may generate the samples that are highly similar to the samples of class $a$ (from the victim's dataset).
Thus, the sensitive information contained in the victim's samples may be leaked to the adversary without the victim's awareness.

The training procedures are summarized as follows.
%\vspace{-0.1 in}

\begin{itemize}[]
	\item [1)]
	\emph{The victim's procedure:}
	\begin{itemize}
		\item [a)]
		The victim downloads the global model from the central server.
		\item [b)]
		The victim uses the global model's parameters to update the local model.
		\item [c)]
		The victim trains the local model using local dataset (with labels $\{a,b\}$).
		\item [d)]
		The victim uploads the updates of the local model to the central server.
	\end{itemize}
	
	\item [2)]
	\emph{The	adversary's procedure:}
	\begin{itemize}
		\item [a)]
		The adversary downloads global model from central server.
		\item [b)]
		The adversary uses the global model's parameters to update the local model.
		\item [c)]
		The adversary copies the global model's parameters to the discriminator.
		\item [d)]
		The adversary trains the generator (discriminator is fixed).
		\item [e)]
		The trained local generator is used to generate some fake samples (similar to samples of class $a$), which are labeled as $c$.
		\item [f)]
		The adversary injects these fake samples into the local dataset.
		\item [g)]
		The adversary trains the local model on local dataset (with labels $\{b,c\}$).
		\item [h)]
		The adversary uploads the updates of the local model to central server.
	\end{itemize}
	%ML model
	\item [3)] Repeat the \emph{victim's procedure} and the \emph{adversary's procedure} by multiple times, then both $\overline{G}$ and $\overline{D}$ can be well trained.
	\item [4)] Finally, the adversary can feed a random noise to $\overline{G}$, which can generate an image (with the target label $a$).
\end{itemize}

Note that the proposed \attname attack can also work in case there are multiple clients.
%
%The pseudo-code of the algorithm is shown in the Appendix \ref{sect:attalg} (Algorithm \ref{CCG-Algorithm}). 
%
The experimental evaluation for \attname attack is presented in Section \ref{sec:Experiment}.
%The experimental evaluation for \attname attack is presented in Section \ref{sec:Experiment}.
%
We find that \attname attack can indeed break the ability of federated learning to protect client data privacy.
Moreover, \attname attack is successful both in FedAvg algorithm or other state-of-the-art FL algorithms, such as FedProx\cite{li2020federated}, SCAFFOLD\cite{karimireddy2020scaffold}, and MOON\cite{li2021model}.
Thus, it is imperative to improve the traditional FL scheme's security performance to mitigate \attname attack.

\section{\name Overview}\label{sec:system}
In this section, we introduce the overview of \name, which can be used to mitigate \attname attack in FL.

\subsection{Key Ideas in Mitigating \attname Attack}

\begin{figure*}[h]
	\centering
	\subfigure[In \name, an adversary  can influence about $1/K$ weights of the ensemble global model. The ensemble global model is viewed as a standalone model.]{\includegraphics[width=0.47\textwidth]{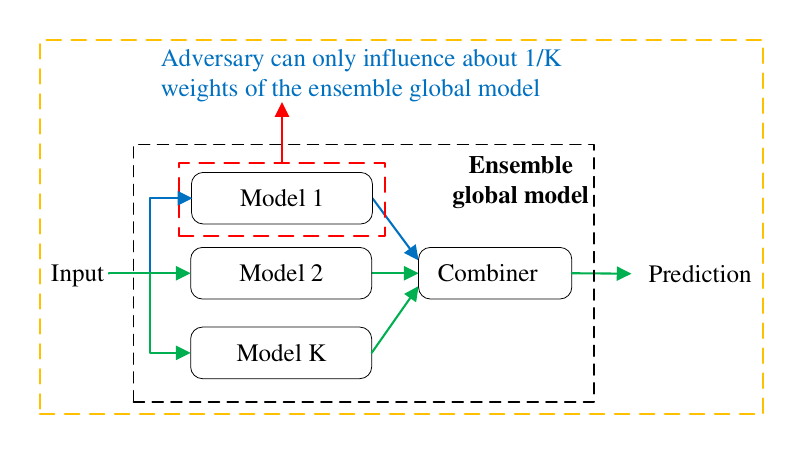}
		\label{subfig2}
	}\hspace{2mm}
	\subfigure[In FedAvg, an adversary can influence all the weights of the averaged global model due to the average operation. ]{\includegraphics[width=0.47\textwidth]{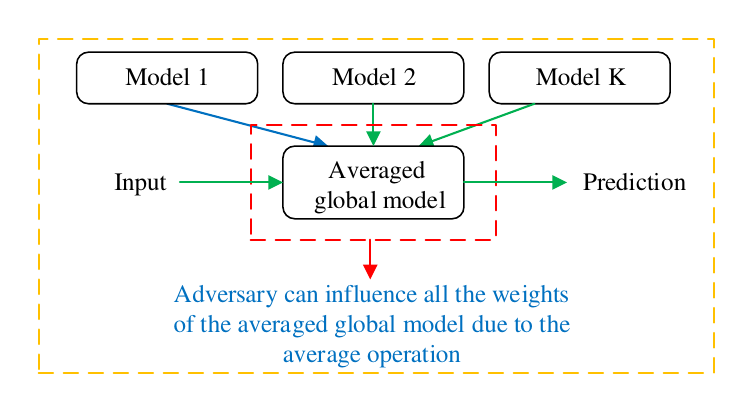}
		\label{subfig1}
	}
	\caption{\centering{\name v.s. FedAvg.}}
	\label{fig:FL_Ensemble}
	%\vspace{0.2in}
	
\end{figure*}

To mitigate \attname attack in FL, we propose \name, which exploits the ensemble learning strategy.
Compared with FedAvg, in \name, the adversary has a weaker control capability over the global model, so \name can mitigate \attname attack.
The adversary can manipulate its local model to directly influence about $1/K$ weights of the ensemble global model in \name, whereas the adversary can directly influence all the weights of the global model in FedAvg.
The reasons are analyzed as follows.

Fig. \ref{subfig2} shows how \name works.
The ensemble operation is finished by a certain combiner.
Given multiple models from clients and input, the server combines the outputs of each model and obtains the final output.
The $K$ local models and the combiner can be viewed as a standalone ensemble global model.
Each client contributes to about $1/K$ weights of the ensemble global model.
Therefore, an adversary can manipulate its local model to directly influence about $1/K$ weights of the ensemble global model.

Fig. \ref{subfig1} shows how the traditional FedAvg works.
The central server averages $K$ local models and then gets an averaged global model.
The averaged global model has the same structure as the local models.
Hence, an adversary can manipulate its local model to directly influence all the weights of the averaged global model.

In a nutshell, because \name significantly reduces the adversary's direct influence over the global model, \name can mitigate \attname attack.

\subsection{\name System Model}

\begin{figure*}[h]
	\centering
	\includegraphics[width=1.8\columnwidth]{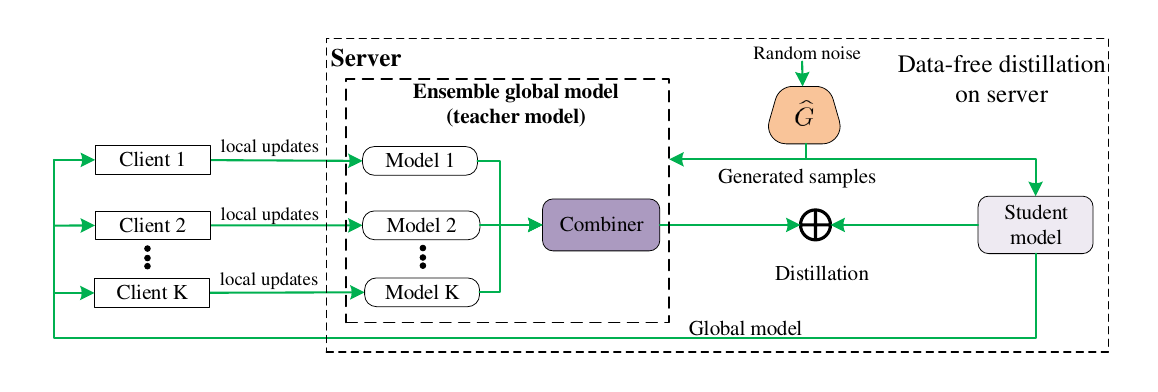}
	%\vspace{-0.2in}
	\caption{\centering{The system model of \name.}}
	\label{fig:FEDED}
\end{figure*}

\name system model is depicted in Fig \ref{fig:FEDED}.
In \name, the server stores each client's local model in the previous round.
After receiving the updates from clients, the server can get its latest model.
The server combines each client's latest model (as shown in Fig. \ref{subfig2}) and gets the ensemble global model.
Then, the server adopts the data-free knowledge distillation technique to distill the knowledge from the ensemble global model into the student model, which is described in Section \ref{sec:data-free} in detail.
The student model is set to be the same size as the original model, so the model expansion problem can be avoided. 
Next, the server sends the compressed student model to each client for the next round of training.
After multiple rounds of training, the student model is well trained and used as the final ML model.

\begin{algorithm}[ht]
	%\small
	\caption{Overall \name Training Algorithm}\label{alg:DFED}
	\begin{algorithmic}[1]
		\Require The $K$ clients are indexed by $k$; $\eta$ is learning rate; $\mathcal{D}_{k}$ is local dataset; $\boldsymbol{w}$ is global model parameters; $E$ is the number of local epochs; $M^{k}$ is clients' latest model; $M_{S}$ is student model.
		\Ensure The global model parameter $\boldsymbol{w}$.
		
		\Procedure{Server executes:}{}
		\State Randomly initialize $\boldsymbol{w}_{0}$\;
		\For{each round $ t=0,1,...$}
		%$ S_{t} $ $\leftarrow$ randomly select $C$-fraction of $K$ clients\;
		\For{each client  $k $ \textbf{in parallel}}
		\State $\boldsymbol{w}_{t+1}^{k} \leftarrow$ ClientUpdate$(k,\boldsymbol{w}_{t})$\;	
		\State $M^k \leftarrow \boldsymbol{w}_{t+1}^{k}$\;
		\EndFor
		\State $M_{S}\leftarrow$\textbf{DataFreeKD$(\{\boldsymbol{M}^{k}\}_{k=1}^{K})$} $\quad$ // invoke Algorithm \ref{alg:datafree}    
		\State $\boldsymbol{w}_{t+1} \leftarrow M_{S}$\;
		\EndFor
		\EndProcedure
		
		\Procedure{ClientUpdate}{$k,\boldsymbol{w}$}: $\quad$ //
		run on client $k$                 %$Run on client
		\For{each local epoch $j=1,2,... , E$}
		\For{each batch $ b $ $\in$ $\mathcal{D}_{k}$}
		\State $ \boldsymbol{w} \leftarrow \boldsymbol{w}-\eta \cdot \nabla Loss(\boldsymbol{w};b)$\;
		\EndFor
		\EndFor
		\State \Return $\boldsymbol{w}$ to server;
		\EndProcedure
	\end{algorithmic}
\end{algorithm}

%\subsection{Overall \name Training Algorithm}\label{sec:algorithm}

Algorithm \ref{alg:DFED} shows the overall \name training algorithm.
In Algorithm \ref{alg:DFED}, $Loss$ denotes cross-entropy loss function.
The server is responsible for data-free knowledge distillation (line 8).
For ClientUpdate (lines 13-17), \name is the same as FedAvg.
Therefore, \name does not require any protocol modification from the client-side.
During the server execution, the server distributes the latest global model to all clients and collects each client's updated latest local model (lines 5-6).
Then, the server utilizes DataFreeKD (line 8) to distill the knowledge from the ensemble global model (combined with each client's latest local model) into the student model (treated as the global model for the next round).

\subsection{Ensemble Strategy}
Ensemble learning is the process of ML that combines the predictions from multiple models for getting better prediction performance \cite{opitz1999popular, rokach2010ensemble}.
To allow distillation of knowledge from the ensemble global model to a student model. 
%The the soft labels (i.e., the output of the XXX) that the ensemble global model predicts for an input $\boldsymbol{x}$ should be computed.
The logit output of the ensemble global model for an input $\boldsymbol{x}$ should be computed.
\name applies the following commonly used ensemble strategy to compute the output.
Given an input $x$, the logit output of the ensemble global model is represented by $M_{T}(\boldsymbol{x})$.
$M_{T}(\boldsymbol{x})$ is given by 
\begin{equation}
	%	\begin{aligned}
		M_{T}(\boldsymbol{x}) = \dfrac{1}{K} \sum_{k=1}^{K}M^k(\boldsymbol{x}),
		%	\end{aligned}
\end{equation}
where $M^k(x)$ is the logit output of the $k$th local model for the input $\boldsymbol{x}$.

\noindent
\textbf{Model Heterogeneity Accommodation.}
Compared with FedAvg, \name can be used for dealing with another problem: it offers model heterogeneity accommodation. 
Even if the local models submitted from clients are heterogeneous with different network depths, these models can still be combined and compressed at the server. 
Comparing with the ensemble distillation scheme in FedDF \cite{lin2020ensemble}, \name has one biggest advantage that the central server does not need an unlabeled auxiliary dataset.
Note that this paper mainly discusses the security performance of \name, so the experimental verification of model heterogeneity is not presented in this paper.

\section{Data-free Knowledge Distillation}\label{sec:data-free}

Directly using the ensemble strategy has some issues. 
First, if the ensemble global model is viewed as a standalone model and it is sent to each client, then the model expands in each server-client training round. 
Second, if the server directly sends the ensemble global model to each client for the next round of training, the other client's local models are leaked to all clients. 
This may lead to security concerns. 

To address the issues incurred by the ensemble strategy, \name exploits knowledge distillation technique.
This technique enables the server to distill the knowledge of the ensemble global model into a compressed student network (with the same size as clients' local model).
Then, the compressed student model is sent to clients for the next round of training.
Therefore, the model expansion issue is solved. 
Besides, each client's local models are merged into a single compressed model, so the model leakage issue is effectively alleviated.
In the following, the knowledge distillation technique developed in \name is introduced in details.

\subsection{Knowledge Distillation}
Knowledge distillation (KD) is proposed to transfer knowledge from a larger teacher network to a smaller student network \cite{hinton2015distilling}.
\name utilizes KD to extract knowledge in the larger ensemble global model to the smaller student model.
Let $M_{T}$ denote the obtained ensemble global model (teacher) in server in one training round.
Let $M_{S}$ denote the student model, and $\{\boldsymbol{x^i}\}_{i=1}^{n}$ denote a mini-batches of input samples.
The student model can be optimized using the following loss function:
\begin{equation}
	\begin{aligned}
		%\mathcal{L}_{KD} = \dfrac{1}{n}\sum_{i}\mathcal{H}_{cross}(y_{t}, y_{s}),
		\mathcal{L}_{KD} =\dfrac{1}{n}\sum_{i}\mathcal{H}_{cross}(\boldsymbol{y}_{T}^{i}, \boldsymbol{y}_{S}^{i}),
	\end{aligned}
\end{equation}
where $\mathcal{H}_{cross}$ denotes cross-entropy loss, $\boldsymbol{y}_{T}^{i} = M_{T}(\boldsymbol{x}^{i})$ is the logit outputs of the ensemble global model, $\boldsymbol{y}_{S}^{i} = M_{S}(\boldsymbol{x}^{i})$ is the logit outputs of the student model, and $\boldsymbol{x}^{i}$ denotes the training data.

\subsection{GANs for Generating Training Samples}

General knowledge distillation requires a training dataset to ensure its accuracy. However, the server is prohibited to access the clients’ training data in FL. If random data is used in KD, the distilled student model has significant performance degradation compared with the teacher model.

Although data-free KD schemes are not novel to the federated learning community, they are not very useful to mitigate \attname attack.
FedGen \cite{zhu2021data} trains a lightweight generator to generate simulation feature and broadcasts it to clients to guide local training.
FedFTG \cite{zhang2022fine} learns a generator to explore the input space of local models to relieve the issue of direct model aggregation.
Both of the two methods aim to extract some useful information from the uploaded updates from clients to guide the training process, but, they did not make an effort to defense against the adversary in \attname attack.
In \name, we adopt the ensemble strategy to mitigate \attname attack.
Thus, we need to design a special data-free KD scheme.
To address this issue, we design a data-free KD scheme used in \name, which is inspired by \cite{chen2019data}.
The designed data-free KD scheme trains a generator $\widehat{G}$ to generate training samples that can facilitate the transfer of knowledge from the teacher network to the student network.
The generator $\widehat{G}$ is trained based on the following well-designed loss functions. 

\noindent
\textbf{One-hot Loss.}
%
%The considered FL task is image classification.
%
In FL, each client uses cross-entropy loss to train the local model.
The cross-entropy decreases as the predicted probability moves close to the ground-truth label. 
Specifically, after \textit{softmax} calculation in the well-trained ensemble global model, only one entry in the output vector is close to 1, and all other entries are close to 0.
\name uses the one-hot loss to guide $\widehat{G}$ to generate samples that can be used to imitate the distribution of training data. 
Therefore, the generated samples can help to transfer knowledge in KD.
For real training samples, the outputs of the teacher network should be classified into one particular category with a higher probability.
Similarly, the outputs of the ensemble global model should be close to one-hot vectors when inputting the generated samples.
Let $\{\boldsymbol{z^i}\}_{i=1}^{n}$ denote a mini-batches of random variables.
Accordingly, the one-hot loss $\mathcal{L}_{oh}$ is designed as 
\begin{equation}
	\begin{aligned}
		\mathcal{L}_{oh} = \dfrac{1}{n}\sum_{i}\mathcal{H}_{cross}(\boldsymbol{y}_{T}^{i},\boldsymbol{t}^{i}),
	\end{aligned}
\end{equation}
where $\boldsymbol{y}_{T}^{i} = M_T({x}^{i})$, $\boldsymbol{x}^{i} = \widehat{G}(\boldsymbol{z}^{i})$, and $\boldsymbol{t}^{i} = \arg\max\limits_{j}(\boldsymbol{y}_{T}^{i})_{j}$.

\noindent
\textbf{Information Entropy Loss.}
In KD, the number of samples in each class is usually balanced.
Hence, \name adopts the information entropy loss to encourage that $\widehat{G}$ balances the number of the generated samples in each class.
The information entropy loss $\mathcal{L}_{ie}$ is defined as
\begin{equation}
	\begin{aligned}
		\mathcal{L}_{ie} =-\mathcal{H}_{info}(\dfrac{1}{n}\sum_{i}\boldsymbol{y}_{T}^{i}),
	\end{aligned}
\end{equation}
where $\dfrac{1}{n}\sum_{i}\boldsymbol{y}_{T}^{i}$ denotes the frequency distribution of each class of generated samples.
Minimizing $\mathcal{L}_{ie}$ means maximizing the information entropy, which can enforce generating samples with roughly equal numbers in each class.

\noindent
\textbf{Total Loss.}
By combining the one-hot loss and information entropy loss, the total loss function is given by
\begin{equation}
	\begin{aligned}
		\mathcal{L}_{total} = \mathcal{L}_{oh} + \beta\mathcal{L}_{ie},
	\end{aligned}
	\label{fun:TotalLoss}
\end{equation}
where $\beta$ is the hyper-parameter balancing both $\mathcal{L}_{oh}$ and $\mathcal{L}_{ie}$.
Through minimizing $\mathcal{L}_{total}$, $\widehat{G}$ can generate useful samples for KD.

\begin{algorithm}[ht]
	\caption{Data-free Knowledge Distillation}\label{alg:datafree}
	\begin{algorithmic}[1]
		\Require The $K$ clients are indexed by $k$; $M^{k}$ is the $k$th client's latest model; $M_{S}$ is student model.
		\Ensure The student model $M_{S}$ 
		
		\Function{DataFreeKD}{$\{\boldsymbol{M}^{k}\}_{k=1}^{K}$}
		\State Randomly initialize $ M_{S}$ and $\widehat{G}$;
		\For {each epoch $j=1,2..$}  
		\State \textit{/* Phase $\uppercase\expandafter{\romannumeral1}$  */}  
		\State Draw random noise variables $\{\boldsymbol{z^i}\}_{i=1}^{n}$\;
		\State Use $\widehat{G}$ to generate samples $\boldsymbol{x}^{i}\leftarrow\widehat{G}({\boldsymbol{z}}^{i})$\;
		\State \textit{/* Phase $\uppercase\expandafter{\romannumeral2}$  */}  
		\State The ensemble global model input $\boldsymbol{x}^{i}$ : $\boldsymbol{y}_{T}^{i}\leftarrow \dfrac{1}{K} \sum_{k=1}^{K}M^k(\boldsymbol{x}^{i})$\;
		\State Calculate $\mathcal{L}_{total}$ loss\;
		\State Update weights in $\widehat{G}$ to minimize $\mathcal{L}_{total}$\;
		\State \textit{/* Phase $\uppercase\expandafter{\romannumeral3}$  */} 
		\State The student model input $\boldsymbol{x}^{i}$ : $\boldsymbol{y}_{S}^{i}\leftarrow M_{S}(\boldsymbol{x}^{i})$\;
		\State Calculate $ \mathcal{L}_{KD} $ loss\;
		\State Update weights in $M_{S}$ to minimize $ \mathcal{L}_{KD} $\;
		\EndFor
		\EndFunction
	\end{algorithmic}
\end{algorithm}

\noindent
\textbf{Remarks.}
The generated samples by $\widehat{G}$ are used to extract knowledge from the ensemble global model to the student model.
They may not be visually similar to the original training samples. 
Because \name uses the generated samples training KD without accessing the private training data, the clients' local data privacy can be protected.

\subsection{Data-free Knowledge Distillation Algorithm}
%See \textbf{DataFreeKD} in Algorithm X.
%
Data-free knowledge distillation can be divided into three phases, which are shown in Algorithm \ref{alg:datafree}.
In Phase $\uppercase\expandafter{\romannumeral1}$, \name draws a set of random noise variables, and then $\widehat{G}$ takes random noise variables as input and generates a group of samples.
In Phase $\uppercase\expandafter{\romannumeral2}$, the ensemble global model (teacher) takes generated samples as input and then calculates $\mathcal{L}_{total}$.
\name updates weights in $\widehat{G}$ using the gradients calculated by the back-propagation of $\mathcal{L}_{total}$.
In Phase $\uppercase\expandafter{\romannumeral3}$, the student model takes generated samples as input and then calculates $\mathcal{L}_{KD}$.
\name updates weights in the student model using the gradients calculated by the back-propagation of $\mathcal{L}_{KD}$.

\section{EXPERIMENTS}\label{sec:Experiment}

In this section, we test the \attname attack performance, and evaluate the effectiveness of \name.
We test the performance of \attname attack and the performance of \name on three datasets, respectively. 
The description of datasets, the distribution of client data, neural network structure, and the setting of parameters can be found in Appendix \ref{sect:appendix}. %(see supplementary file). 

\subsection{Evaluation of \attname Attack}

\begin{figure}[hb]
	\centering
	\includegraphics[width=0.95\columnwidth]{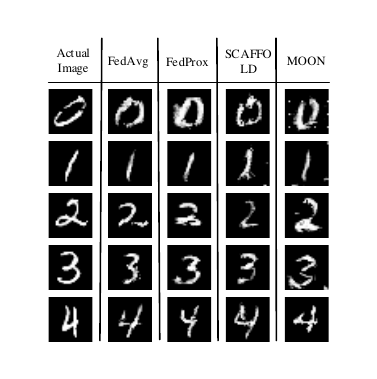}
	%\vspace{-0.15in}
	\caption{The results of \attname attack on FedAvg and other FL variants. The first column is the real dataset images. The rest columns are the reconstructed images by \attname attack on different FL variants.}
	\label{fig:compare}
\end{figure}

\noindent
\textbf{\attname Attack Evaluation on Different FL Algorithms.}
To demonstrate whether \attname attack is effective against the state-of-the-art federated learning algorithm, we apply \attname attack on several FL variants including FedAvg, FedProx \cite{li2020federated}, SCAFFOLD \cite{karimireddy2020scaffold}, and MOON \cite{li2021model}.
The results are shown in Fig. \ref{fig:compare}, the first column is the real images from victims, and the other columns show the reconstructed images by \attname attack against above FL algorithms.
None of above FL algorithms can defend against \attname attack reconstructing visually recognizable images.
In addition, we quantitatively evaluate the similarity between the reconstructed images and target images using the following metrics: (1) mean square error (MSE $\downarrow$); (2) structural similarity index (SSIM $\uparrow$) \cite{wang2004image}; (3) peak signal-to-noise ratio (PSNR $\uparrow$) \cite{hore2010image}; (4) learned perceptual image patch similarity (LPIPS $\downarrow$) \cite{zhang2018unreasonable}.
	Note that $\downarrow$ indicates the lower value of the metric the higher reconstructed image quality, while $\uparrow$ represents the higher value of the metric the higher reconstructed image quality.
Table \ref{fig:quality} shows the similarity between the reconstructed images by \attname attack and the real images when applying different FL algorithms.
%
%Note that the lower the similarity, the higher MSE and LPIPS scores, and the lower the PSNR and SSIM scores.
%
When \name is adopted, the similarity is the lowest.
This prove that the proposed \name can resist against \attname attack reconstructing high quality images.

\begin{table*}[h]
	\centering
   \normalsize
	\begin{tabular}{ccccc}
        
		\toprule
		Algorithm & MSE$\downarrow$ & PSNR$\uparrow$ & SSIM$\uparrow$ & LPIPS$\downarrow$ \\
		\toprule
		FedAvg & 0.0587 & 12.3081 & 0.4893 & 0.1653 \\
		FedProx & 0.0638 & 11.9494 & 0.4487 & 0.1616 \\
		SCAFFOLD & 0.0643 & 11.9170 & 0.5315 & 0.1819 \\
		MOON & 0.0857 & 10.6656 & 0.4119 & 0.2581 \\
		\textbf{\name} & \textbf{0.2275} & \textbf{4.8416} & \textbf{0.0438} & \textbf{0.6357} \\
		\toprule
	\end{tabular}
	\caption{The similarity between the reconstructed images and the real images when applying different FL algorithms. Note that $\downarrow$ indicates the lower value of the metric the higher reconstructed image quality, while $\uparrow$ represents the higher value of the metric the higher reconstructed image quality.}
	\label{fig:quality}
\end{table*}

\noindent\textbf{\attname Attack Evaluation on Different FL Parameter Settings.}
In federated learning, the server randomly selects a part of clients in each round, and aggregates the updates uploaded by these clients.
So, the number of clients $K$ and the fraction of clients $C$ will have an impact on \attname attack.
Table \ref{tab:round} shows the communication rounds required for \attname attack to successfully reconstruct visually recognizable images under different parameter settings.
	We have three observations. 
	First, a smaller $C$ makes \attname attack require larger communication rounds when $K = 10$. 
	This is because a smaller $C$ means the adversary has a lower probability to participate in the training process.
	Second, a larger $K$ makes \attname attack require larger communication rounds when $C = 1$.
	Third, \attname attack cannot successfully reconstruct images when the number of selected clients (i.e., $K \times C$) is large enough per round (e.g., $K = 200$, $C = 1.0$ or $K = 200$, $C = 0.5$).
	Another example is that the required communication rounds in the case ($K = 100, C = 1.0$) is higher than in the case ($K = 100, C = 0.1$).
	This is because too many clients are selected per round, thus, the adversary's adversarial influence becomes trivial after the average update in FL.
	Therefore, the required communication round does not strictly increase as $K$ increases or strictly increases as $C$ decreases.
	Our experiments demonstrate that these two factors ($K$ and $C$) intertwined to influence the performance of \attname attack.

\begin{table*}[h]
	\centering
    \normalsize
	%\resizebox{\linewidth}{!}{
		\begin{tabular}{c@{\extracolsep{6pt}}cccc}
			\toprule
			\multirow{2}{*}{Num of Clients} & \multicolumn{4}{c}{Fraction of Clients}            \\ 
			\cline{2-5}
			& $1.0$ & $0.5$ & $0.3$ & $0.1$            \\ \toprule
			$10$                   & \textbf{20} & 50 & 95 & 149 \\
			$50$                   & 77 & 89  & 72 & 211 \\
			$100$                  & \textbf{393} & 120 & 110 & 188 \\
			$200$    & \XSolidBrush & \XSolidBrush & 198 & 320 \\
			
			\toprule
		\end{tabular}
		%}
	\caption{Communication rounds comparison of different FL parameter settings under \attname attack. \XSolidBrush means that \attname attack cannot successfully reconstruct visually recognizable images.}
	\label{tab:round}
\end{table*}

\noindent\textbf{\attname Attack Evaluation on Different Network Structures.}
We evaluate how different network structures defense \attname attack in Table \ref{tab:difmodels}.
We perform FedAvg on six different network structures and execute the \attname attack simultaneously.
We observe that the \attname attack works well when training network structures, i.e., successfully reconstructing visually recognizable images.
This means different network architectures can not defense \attname attack.

\begin{table*}[h]
	\center
   \normalsize
	\begin{tabular}{|c|c|c|c|c|c|}\hline
		% \multicolumn{1}{|c|}{\multirow{2}{*}{aaaa}} & \multicolumn{3}{c|}{bbbb}\\ \cline{2-4}
		%  & cccc  & dddd & eeee \\ \hline
		\multirow{1}{*}{\textbf{Nets}} & MLP &LeNet & GoogLeNet\\ \hline
		\multirow{1}{*}{\textbf{Succeed}} & \Checkmark & \Checkmark & \Checkmark  \\ \hline
		\multirow{1}{*}{\textbf{Nets}} & VGGNet & AlexNet & ResNet18\\ \hline
		\multirow{1}{*}{\textbf{Succeed}} & \Checkmark & \Checkmark & \Checkmark  \\ \hline
	\end{tabular}
	
	\caption{Training different ML models using FedAvg under \attname attack. \Checkmark means the reconstructed images by \attname attack are visually recognizable.}
	
	\label{tab:difmodels}
	
\end{table*}

\begin{figure}[h]
	\centering
	\includegraphics[width=0.99\columnwidth]{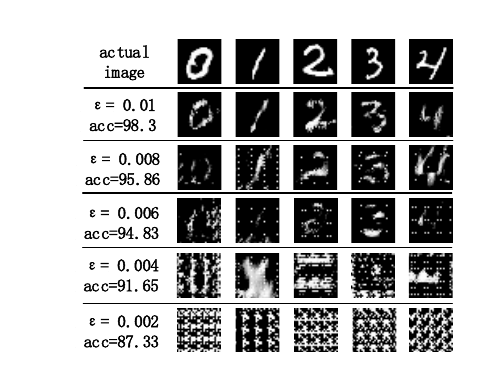}
	%\vspace{-0.15in}
	\caption{The results of \attname attack on FedAvg with differential privacy enabled.}
	\label{fig:dp}
\end{figure}

\noindent
\textbf{\attname Attack Evaluation on FedAvg with Differential Privacy Enabled.}
We conduct \attname attack experiment on FedAvg algorithm with differential privacy (proposed in \cite{abadi2016deep}) enabled.
For the training process of federated learning under \attname attack, we keep the same experiment settings with the normal FedAvg algorithm.
In addition, we define a privacy budget $ \epsilon$ and apply the gaussian mechanism to add noise to the updates before sharing them.
The results are shown in Fig. \ref{fig:dp}.
The smaller privacy budget $ \epsilon$ means higher security but leads to lower accuracy of the global model. 
We find that as the privacy budget $\epsilon$ decreases, the images reconstructed by \attname attack are getting visually unrecognizable, but the accuracy of the trained global model is also decreasing.
Although the accuracy of the model does not drop significantly when the privacy budget $\epsilon = 0.1$, the \attname attack can still successfully reconstruct visually recognizable images.
When the reconstructed images are completely unrecognizable ($\epsilon = 0.004$), the accuracy of the trained global model is only 91.65\%, that is 98.5\% (in the next subsection) in \name.
This demonstrates that adopting differential privacy in FL can indeed mitigate \attname attack, but it will incur a much larger accuracy degradation.

\noindent
\textbf{The Effectiveness of Various Defense Strategies.}
According to previous studies, three defense methods are evaluated: 
(1) Additive Noise \cite{zhu2019deep}: gaussian noise $\epsilon \sim \mathcal{N}(0, \sigma^2\boldsymbol{I})$ is injected into the updates with $\sigma$ range from 0.05 to 0.3; 
(2) Gradient Clipping \cite{geyer2017differentially}: clip the value of the updates with the bound from 4 to 0.1; 
(3) Gradient Spasification \cite{lin2017deep}: perform different levels of sparsities range from 10\% to 90\%.
We apply these methods to the uploaded local model updates before sharing them.
As shown in Table \ref{tab:defense}, when the $\sigma = 0.05$  or $\sigma = 0.1$, the additive noise does not successfully defense \attname attack.
\attname attack fails to perform only when $\sigma >= 0.2$ and the added noise starts to significantly affect model accuracy.
Compared with \name, additive noise requires a larger drop in accuracy to successfully defense against \attname attack.
The gradient clipping can also successfully resist \attname attack and only bring a slight accuracy degradation.
The accuracy of the trained global model by \name in IID case is nearly 98.5\% (in the next subsection), and that by gradient clipping is approaching 97\%.
In addition, we evaluate different levels of sparsity on updates (range from 10\% to 90\%).
Although the higher sparsity does not lead to a noticeable drop in accuracy, it has almost no effects against \attname attack.

\begin{table*}[htbp]
	\centering
   \normalsize
	\begin{tabular}{ccccccc}
		\toprule
		\multirow{4}{*}{Additive Noise} & $\sigma$ & 0.05 & 0.1 & 0.2 & 0.3  \\ \cline{2-6}
		& Accuracy & 98.7\% & 96.7\% & 93.8\% & 89.8\%  \\ \cline{2-6}
		& Defendability & \XSolidBrush    & \XSolidBrush   & \Checkmark  & \Checkmark\\ \hline
		\multirow{4}{*}{Gradient Clipping} &  bound & 4 & 1 & 0.5 & 0.1  \\ \cline{2-6}
		& Accuracy & 99.1\% & 98.6\% & 98.4\% & 97\%  \\ \cline{2-6}
		& Defendability &  \XSolidBrush     & \XSolidBrush     & \XSolidBrush    & \Checkmark \\ \hline
		\multirow{4}{*}{Gradient Sparsification} & sparsity & 10\% & 30\% & 60\% & 90\%  \\ \cline{2-6}
		& Accuracy & 99.1\% & 99.2\% & 99.1\% & 99.0\%  \\ \cline{2-6}
		& Defendability & \XSolidBrush   & \XSolidBrush    & \XSolidBrush   & \XSolidBrush    \\ 
		\toprule
	\end{tabular}
	\caption{The trade-off between accuracy and defendability with various defense methods under \attname attack. \Checkmark means it successfully defenses against \attname attack while \XSolidBrush means fails to defense \attname attack (whether the reconstructed samples are visually recognizable). The results are evaluated on MNIST with the IID case (FedAvg).}
	\label{tab:defense}
\end{table*}

\subsection{Evaluation of \name}

\begin{figure}[h]
	\centering
	\includegraphics[width=0.99\columnwidth]{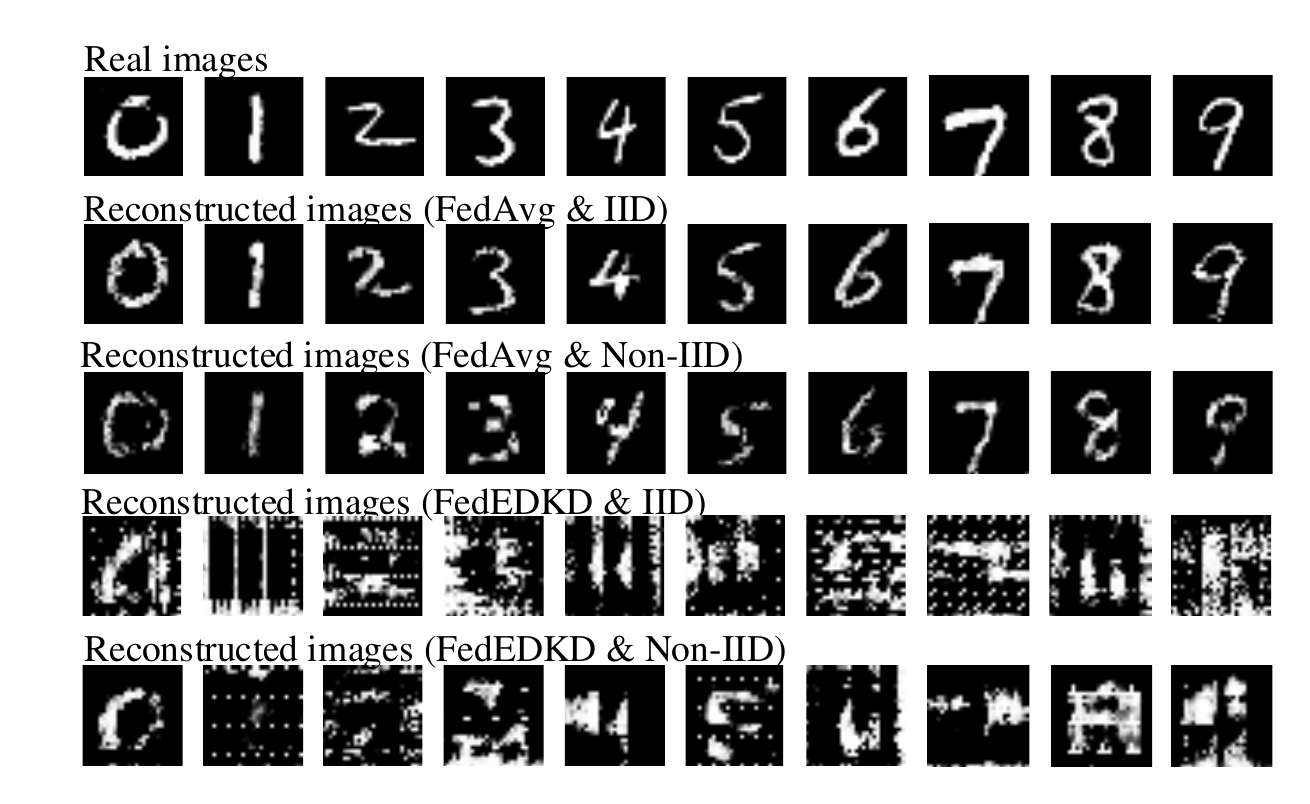}
	%\vspace{-0.15in}
	\caption{The results of \attname attack on FedAvg and \name. The first row is the real images from victims. The second row is the reconstructed images by \attname attack in the Fedavg $\&$ IID case. The third row is the reconstructed images by \attname attack in the Fedavg $\&$ Non-IID case. The last two rows are the reconstructed images by \attname attack in the \name.}
	\label{fig:FL-EDKD}
\end{figure}
%\subsubsection{\attname Attack Evaluation on FedAvg}
\noindent
\textbf{\attname Attack Evaluation on FedAvg and \name.}
We first conduct \attname attack experiment on MNIST dataset \cite{lecun1998mnist} by varying the target label $a=0,\cdots,9$ to test \attname attack on FedAvg and \name.
The results are shown in Fig. \ref{fig:FL-EDKD}.
We have the following observations, first, in both IID and Non-IID cases, \attname attack enables the adversary to successfully reconstruct visually recognizable images in FedAvg, second, the reconstructed images follow the same distribution as the images of the target label from victims.
To verify the effectiveness of \name defending \attname attack, we repeat the above experiments on \name.
Compared with the results in FedAvg, Fig \ref{fig:FL-EDKD} (the last two rows) shows that most of the images reconstructed by \attname attack are significantly blurred when using \name.
That is, \attname attack can be mitigated by \name.

\begin{figure}[h]
	\centering
	\includegraphics[width=0.99\columnwidth]{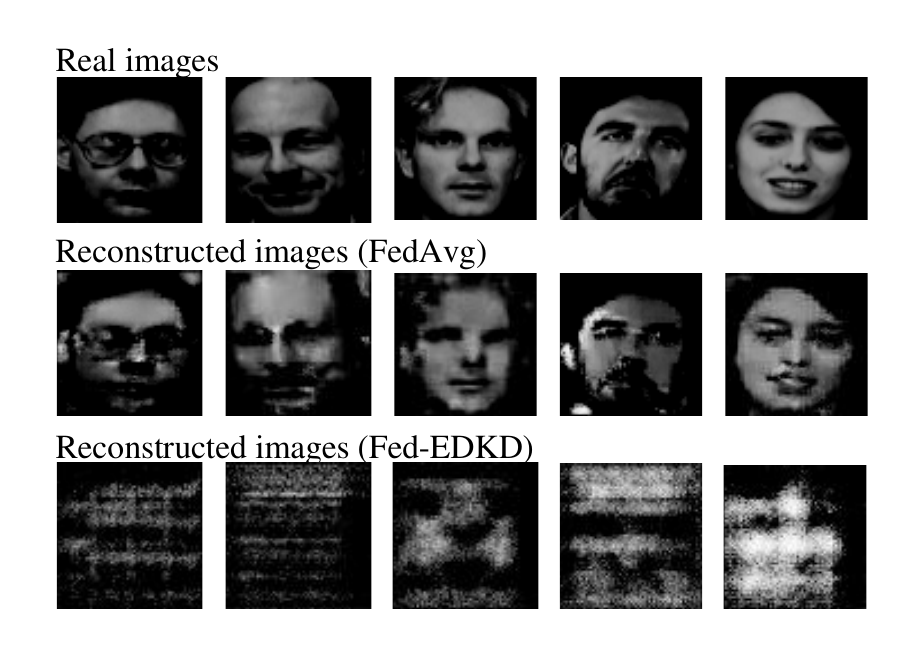}
	%\vspace{-0.2in}
	\caption{The results of \attname attack on FedAvg. The first row is the real images from victims. The second row is the reconstructed images by \attname attack in FedAvg.The third row is the reconstructed images by \attname attack in \name.}
	\label{fig:ATT}
\end{figure}

% \noindent
% \textbf{\attname Attack Evaluation on FedAvg and \name for AT\&T Dataset.}
%
Fig. \ref{fig:ATT} shows the results of \attname attack on FedAvg and \name when adopting AT\&T dataset \cite{samaria1994parameterisation}.
The first row shows some of the real images from the victim (each image is one of the face images from one target class).
The second row is the reconstructed images (subjectively selected the corresponding image that is most similar to the victim's image) by \attname attack in FedAvg case.
The third row is the reconstructed images by \attname attack when \name is used.
We can observe that the reconstructed images by \attname attack in FedAvg contain the same characteristics as the real images from the victim, such as wearing glasses (first image).
Although the second row reconstructed images have a certain distortion compared with the real data, it is easy for ones to infer private information from the reconstructed images.
For the reconstructed images in the third row, almost all features are lost so that they cannot be distinguished what they are.
These confirm that \name can mitigate \attname attack.

\begin{figure}[h]
	\centering
	\includegraphics[width=0.99\columnwidth]{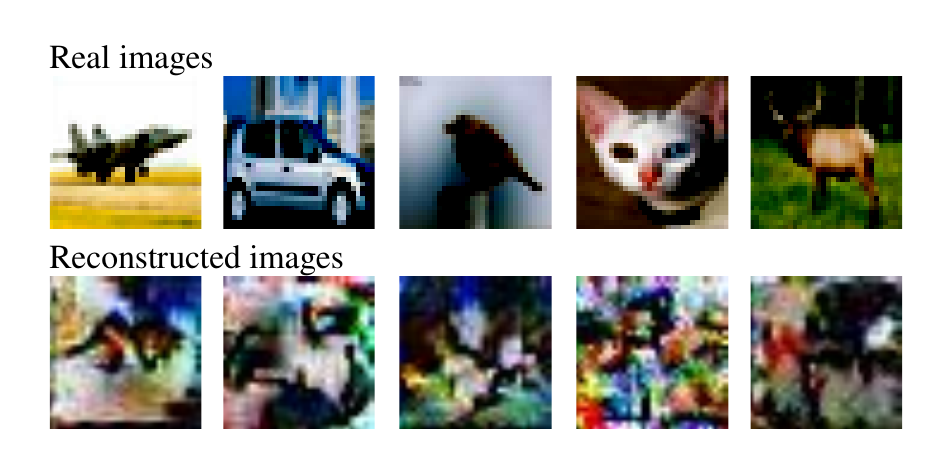}
	%\vspace{-0.2in}
	\caption{The results of \attname attack on FedAvg (CIFAR-10). The first row is the real images from victims. The second row is the reconstructed images by \attname attack in FedAvg.}
	\label{fig:cifar}
\end{figure}

% \noindent
% \textbf{\attname Attack Evaluation on FedAvg for CIFAR10 Dataset.}
Fig. \ref{fig:cifar} shows the results of \attname attack on FedAvg using CIFAR-10 dataset \cite{krizhevsky2009learning}.
As shown in Fig. \ref{fig:cifar}, the original images are hard to be inferred from the reconstructed images. 
Therefore, \attname attack has a poor performance on CIFAR-10, in which the images of the target class distribute much larger to each other.
Because \attname attack has a poor performance on CIFAR-10, we do not need to do experiments to test the performance of \attname attack on \name. 

\begin{figure}[h]
	\begin{center}
		\subfigure[IID distribution.]{
			\label{acc:iid}
			\centering
			\includegraphics[width=2.5 in]{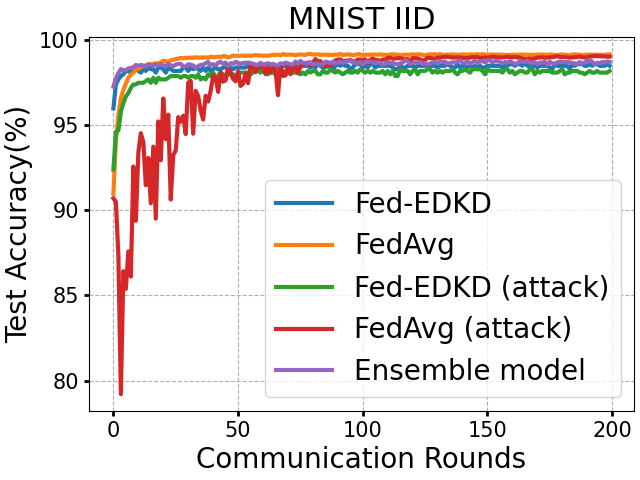}
		}\hspace{0mm}
		\subfigure[Non-IID distribution.]{
			\label{acc:non-iid}
			\centering
			\includegraphics[width=2.5 in]{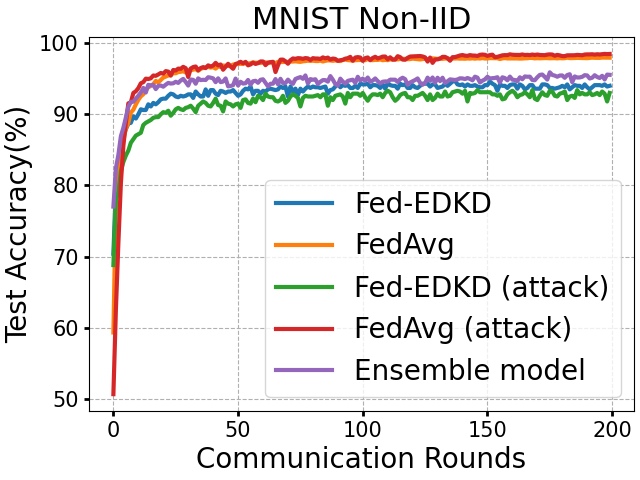}
		}\hspace{0mm}
	\end{center}
	%\vspace{-0.1in}
	\caption{\centering{Test accuracy v.s. communication rounds.}} \label{fig:accuracy}
\end{figure}

%\subsubsection{\name Performance}
\noindent
\textbf{\name Performance.}
We compare the test accuracy of the global model trained by \name and FedAvg in Fig. \ref{fig:accuracy}.
The comparison experiments are conducted on MNIST in IID and Non-IID cases.
The test accuracy of the well-trained global model trained by FedAvg in the IID case is nearly 99.2\%, and that by \name is approaching 98.5\%.
The test accuracy of the well-trained global model trained by FedAvg in the Non-IID case is nearly 97.9\%, and that by \name is approaching 94.7\%.
The test accuracy of the trained ML model is only reduced by 0.7\% in the IID case and is reduced by 3.2\% in the Non-IID case. 
Thus, \name only incurs a slight accuracy degradation on the trained global model compared with FedAvg.

In the design of \attname, the generated fake samples of target class $a$ are labeled with $c$, so, does these fake samples affect the training of the global model?
As shown in Fig. \ref{fig:accuracy}, we find that \attname attack only causes a slight drop in accuracy after the model finally converges.
In the IID case, Fig. \ref{acc:iid} shows that \attname attack causes the model to converge more slowly on FedAvg, but does not have a significant impact on \name.
This also demonstrates that the malicious client’s control capability over the global ML model is significantly reduce in \name.
In the Non-IID case, \ref{acc:non-iid} shows that \attname attack did not have a noticeable impact both on FedAvg and \name.

	In \name, we use cross-entropy loss to perform data-free knowledge distillation scheme.
	To evaluate the loss of knowledge (i.e., model accuracy) during the distillation process, we test the accuracy of ensemble model (teacher model) and compare it with the accuracy of distillation model (student model).
	As shown in Fig. \ref{fig:accuracy}, the accuracy curve of ensemble model(i.e., purple curve) and the accuracy curve of distillation model(i.e., blue curve) are very close.
	The test accuracy of ensemble model in the IID case is nearly 98.82\%, and that of distillation model is approaching 98.5\%.
	Furthermore, the test accuracy of the ensemble model in the Non-IID case is nearly 95.69\%, and that of distillation model is approaching 94.7\%.
	The test accuracy of distillation model is only reduced by no more than 1\%.
	Thus, the operation of data-free knowledge distillation only incurs a slight accuracy degradation.

	In \name, there is no modification on the client-side operation, while the server-side operations are updated.    
	We evaluate the computational costs of client and server on several different FL algorithms, which is depicted in Table \ref{tab:cost}.
	As expected, in \name, the computational cost of a client is nearly the same with FedAvg and SCAFFOLD, because their protocols on the client-side are almost identical.
	Furthermore, the computational cost of the server in \name is much higher than other schemes. 
	The root cause is that the data-free knowledge distillation operation in \name requires more computation overhead. 
	However, the extra computational cost incurred by \name is not a big issue.
	First, since the server (e.g., cloud server) is usually much more powerful in FL. 
	We can assign more computing resources to the server to accelerate its computation in real-world applications if needed. 
	Second, most FL applications are not real-time, so waiting for several minutes to perform one iteration is acceptable. 
	In conclusion, \name works by investing more computations to gain security. 

\begin{table*}[h]
	\centering
	\caption{Computational costs comparison of client and server on different FL algorithms.}
	\label{tab:cost}
	\resizebox{\linewidth}{!}{
		\begin{tabular}{cccccc}
			\toprule
			
			Time (seconds) & FedAvg & FedProx & SCAFFOLD & MOON & \name\\
			\midrule
			client & $2.29 \pm 0.2$  & $3.31 \pm 0.15$ & $2.38 \pm 0.1$ & $16.28 \pm 2.1$ & $2.12 \pm 0.08$\\
			server & $0.0010 \pm 0.0002$ & $0.0011 \pm 0.0002$ & $0.0010 \pm 0.0001$ & $0.0012 \pm 0.0002$ & $65.283 \pm 2.1300$ \\
			\bottomrule
		\end{tabular}
	}
\end{table*}

Comparing the experiments on the above three datasets, we have the following observations.
For the performance of \attname attack, if the samples of the target class distribute closer to each other, then \attname attack reconstructed samples can leak more information to the adversary.
For example, different images with label ``1" (in MNIST dataset) distribute closer to each other, whereas different images with label ``automobile" (in CIFAR-10 dataset) distribute relatively faraway to each other.
Consequently, \attname attack has better performance on MNIST dataset than that on CIFAR-10 dataset. 
	Note that \attname attack only works well on the dataset in which the data distribute close to each other for a given class, but there are already many data types that have such a property. 
	For example, many biometric datasets (e.g., face image datasets, fingerprint datasets, iris datasets, etc) have such a property.
	Thus, the \attname attack indeed imposes serious threats to FL. 
For the performance of \name, our experiments demonstrate that \name is capable of mitigating \attname attack in FL on different datasets. 
The adversary reconstructed samples are highly blurred via using \name, which means \name mitigates the \attname attack. 
Furthermore, \name only incurs a slight performance degradation.
For example, on MINIST dataset, the test accuracy of the trained ML model is only reduced by 0.7\% from 99.2\% (using FedAvg) to 98.5\% (using \name) in the IID case.

\section{Discussions}\label{sec:discuss}
	Some advantages and limitations of \name are discussed in this section. 
	\noindent
	\textbf{Handing Model Heterogeneity.} The proposed \name can be used to handle model heterogeneity in federated learning. 
	If the local models submitted from clients are heterogeneous with different network depths,  these different client models can still be combined and compressed at the server.

	\noindent
	\textbf{Defending Against Poisoning Attacks.}
	\name can mitigate data poisoning attacks in FL. 
	Based on the ensemble strategy used in \name, poisonous updates (incurred by poisoning data) also can be mitigated when the number of malicious clients is limited.

	\noindent
	\textbf{Extra Computational Cost.}
	\name needs more computational cost on the server.
	Since the server (e.g., cloud server) is usually much more powerful than client-side devices, the extra computation cost is acceptable. 
	This observation follows a commonly seen principle: security is not free.  
	\name pays the computational cost in order to gain security in FL.

% Compared with existing FL algorithms, \name has other advances, such as model heterogeneity accommodation and defense against poisoning attacks.
% %
% First, \name offers model heterogeneity accommodation where these client models can still be combined and compressed at the server even if the local models submitted from clients are heterogeneous with different network depths. 
% %
% Second, according to the ensemble strategy in \name, poisonous updates also can be mitigated when the number of malicious clients is limited.
% %
% Note that this paper mainly discusses the security performance of \name, so the experimental verification of model heterogeneity and poisoning attack is not presented in this paper.

%In this study, \name is designed to defend against \attname attack.
%
%In Section \ref{sec:Experiment}, we have evaluated some advances and limitations of \name.
%
%For the advances, \name can effectively mitigate C-GANs attack, while the baseline FL algorithm (i.e., FedAvg) and state-of-the-art FL algorithm(e.g., FedProx, SCAFFOLD, and MOON) neither can prevent privacy leakage under \attname attack.
%
%Besides, the protocol of \name on client-side is the same with FedAvg, and \name can achieve higher performance without sacrificing security compared with several defense strategies (e.g., differential privacy, additive noise, etc).

\section{ CONCLUSION}\label{sec:conclusions}
%
% In this paper, we have tested the \attname attack performance in FL and found conditions under which the attack has good/poor performance.
% %
% We have proposed \name to mitigate \attname attack while only incurring a slight performance degradation of FL.
% %
% \name does not require to revise any protocol from the client-side. 
% %
% Besides, \name can achieve higher performance without sacrificing security compared with several defense methods.
% %
% In summary, \name is suitable to be used in the security-critical FL applications.

	In this paper, we tested the \attname attack performance in FL and found conditions under which the attack has good/poor performance.
	Furthermore, we improve the traditional FL scheme and further propose \name to mitigate \attname attack.
	The main idea of \name is to adopt the ensemble learning scheme in FL to reduce malicious clients' control capability over the global ML model.
	First, the server combines the received client models and obtains an ensemble global model using the ensemble strategy.
	Then, we employ the knowledge distillation technique to transfer the knowledge from the large ensemble model to a smaller compressed student model which has the same size as the original model.
	Therefore, the student model can perform multiple iterations of training between clients and server, which can obtain high performance of the global model.
	Besides, we exploit the data-free knowledge distillation technique to solve the challenge of lacking real training data.
	Specifically, a generator is trained to produce the imitated training data to facilitate the knowledge transfer from the teacher to the student mode.
	Finally, intensive simulations have demonstrated the effectiveness and superiorities of \name. 
	\name has the following impacts. 
	First, \name significantly mitigates \attname attack on two datasets while only incurring a slight accuracy degradation of FL.
	Second, \name can achieve higher performance without sacrificing security compared with several defense strategies (e.g., differential privacy, additive noise, etc).
	Third, the ideas proposed in \name can have broader impacts since they can be used to handle model Heterogeneity and to defend against data poisoning attacks in FL. 
	There are two directions to do future work. 
	First, we plan to study how to use \name to handle heterogeneous FL, in which the local models submitted
	from clients are heterogeneous with different network depths. 
	Second, we plan to empirically measure the performance of \name in the presence of data poisoning attacks.

\section{Acknowledgements}
This work was supported by the National Key R\&D Program of China under Grant 2022YFB3103500, the National Natural Science Foundation of China under Grants 62072062 and U20A20176, Natural Science Foundation of Chongqing under Grant cstc2022ycjh-bgzxm0031, CCF-AFSG Research Fund under Grant RF20220009, and National Science Foundation (Grant no. CNS-2153393).

%
%\section*{Data Availability Statements}
%The datasets generated during and/or analyzed during the current study are available from the corresponding author upon reasonable request.
%
%\section*{Competing Interests}
%The authors have no relevant financial or non-financial interests to disclose.

\begin{appendices}

	\section{Experimental DETAILS}
	\label{sect:appendix}

	\subsection{Experiments on MNIST}
	%MNIST
	
	\noindent
	\textbf{Dataset.} MNIST dataset \cite{lecun1998mnist} consists of 70,000 hand-written digits images from 0 to 9.
	Specifically, the training set has 60,000 images and the testing set has 10,000 images.
	Each image is in grayscale with $28 \times 28$ pixels.
	
	\noindent
	\textbf{Distribution of Client Data.}
	Setting the number of clients $K=10$, we study two different data distributions over clients, i.e., IID and Non-IID.
	For the IID case, we first randomly shuffle the data and evenly divide it into $10$ equal shards, then, assign 1 shard to each client.
	For the adversary, we just remove the data with target label $a$ to ensure the adversary has no data with the target label $a$.
	For the non-IID case, we first sort the data by digit label and separate it into $20$ equal shards, then, assign $2$ shards to each client such that each client only has the images with two digit labels.
	Especially, during the assignment, we have to guarantee that the data with target label $a$ cannot be assigned to the adversary.

	\noindent
	\textbf{Neural Network Structure.}
	The classifier and discriminator have the same network structure.
	In the classifier/discriminator, the network input size is $28\times28\times1$, and the network output size is $1\times10\times1$.
	The classifier/discriminator network consists of convolutional layers and fully connected layers.
	The kernel size in convolutional layers is $5\times5$ with the stride of $1$.
	The adversary's generator $\overline{G}$ is constructed by deconvolutional layers and batch normalization layers.
	In deconvolution layers, the kernel size is $4$, the stride is $ 2 $, and the padding is $ 1 $.
	The generator $\widehat{G}$ for distillation follows \cite{radford2015unsupervised}, and $\widehat{G}$ is constructed by upsampling, convolutional layers, and batch normalization layers.
	The kernel size in convolutional layers is $3\times3$ with the stride of $1$, and the padding is $ 1 $.

	%MNIST
	\noindent
	\textbf{Parameters Setting.}
	(1) \emph{FedAvg}. The number of clients is $K=10$.
	FL task (classification) is optimized using stochastic gradient descent (SGD) with a learning rate of $0.1$.
	The fraction of clients is $C=1$.
	The number of local epochs is $E=1$ and batch size is $20$.
	(2) \emph{\name}. FL task is optimized using SGD with a learning rate of $0.1$.
	The number of local epochs is $E=1$ and batch size is $20$.
	The generator $\widehat{G}$ is optimized using Adam \cite{kingma2014adam} with a learning rate of $0.1$.
	The student model is optimized using Adam with a learning rate of $0.002$.
	For the total loss in Eq. (\ref{fun:TotalLoss}), $\beta$ takes 5.
	For the adversary, $\overline{G}$ is optimized using SGD with a learning rate of $0.0005$.
	
	\subsection{Experiments on CIFAR-10}
	%CIFAR-10
	\noindent
	\textbf{Dataset.} CIFAR-10 dataset \cite{krizhevsky2009learning} consists of 60,000 color images in 10 classes.
	Specifically, there are 50,000 training images and 10,000 testing images.
	Each image is in size with $3 \times 32 \times 32$ pixels.
	
	\noindent
	\textbf{Distribution of Client Data.}
	For CIFAR-10 dataset, we test two-client scenarios where one is the victim and the other is the adversary.
	Each client is assigned 5 classes.
	
	%ATT
	%\subsubsection{Experimental Setup}
	%CIFAR-10
	\noindent
	\textbf{Parameters Setting.}
	(1) \emph{FedAvg.} The number of clients is $K=2$.
	FL task (classification) is optimized using SGD with a learning rate of $0.01$.
	The fraction of clients is $C=1$.
	The number of local epochs is $E=5$ and batch size is $50$.
	
	\noindent
	\textbf{Neural Network Structure.}
	For CIFAR-10 classifier/discriminator, the network structure adopts ResNet18.
	The used generator $\overline{G}$ structure follows the proposed generator in DCGAN \cite{radford2015unsupervised}.

	\subsection{Experiments on AT\&T}
	
	\noindent
	\textbf{Dataset.} AT\&T dataset \cite{samaria1994parameterisation} consists of 400 face images from 40 different persons, namely 10 images per person. 
	Each image is in grayscale with $92 \times 112$ pixels. 
	In this paper, we first resize the face images into $64 \times 64$.
	Then, we randomly divide the AT\&T dataset into a training set and a test set with a ratio of 0.8, 0.2 respectively.
	
	\noindent
	\textbf{Distribution of Client Data.}
	For AT\&T dataset, we test two-client scenarios where one is the victim and the other is the adversary.
	Each client is assigned 20 classes.

	\noindent
	\textbf{Neural Network Structure.}
	For AT\&T classifier/discriminator, the network input size is $64\times64\times1$, and the network output size is $1\times40\times1$.
	The AT\&T classifier/discriminator network consists of convolutional layers and fully connected layers.
	The kernel size in convolutional layers is $5\times5$ with the stride of $1$.
	The generator $\overline{G}$ and $\widehat{G}$ for AT\&T are similar to MNIST generator with the same kernel size.
	%

	%ATT
	\noindent
	\textbf{Parameters Setting.}
	(1) \emph{FedAvg.} The number of clients is $K=2$.
	FL task (classification) is optimized using SGD with a learning rate of $0.1$.
	The fraction of clients is $C=1$.
	The number of local epochs is $E=10$ and batch size is $5$.
	(2) \emph{\name}. FL task is optimized using SGD with a learning rate of $0.1$.
	The number of local epochs is $E=10$ and batch size is $ 5 $.
	The generator $\widehat{G}$ is optimized using Adam with a learning rate of $0.1$.
	The student model is optimized using Adam with a learning rate of $0.002$.
	For the total loss in Eq. (\ref{fun:TotalLoss}), $\beta$ is 5.
	For the adversary, $\overline{G}$ is optimized using SGD with a learning rate of $0.0001$.

\end{appendices}





\end{document}